\def\year{2022}\relax
\documentclass[letterpaper]{article} 
\usepackage{aaai22}  
\usepackage{times}  
\usepackage{helvet}  
\usepackage{courier}  
\usepackage[hyphens]{url}  
\usepackage{graphicx} 
\usepackage{natbib}  
\usepackage{caption} 
\DeclareCaptionStyle{ruled}{labelfont=normalfont,labelsep=colon,strut=off} 
\usepackage{algorithm}
\usepackage{algorithmic}

\usepackage{newfloat}
\usepackage{listings}
\floatstyle{ruled}
\newfloat{listing}{tb}{lst}{}
\floatname{listing}{Listing}
\usepackage{latexsym}
\usepackage{amssymb}
\usepackage[utf8]{inputenc}
\usepackage{microtype}
\usepackage{amsmath}
\usepackage{array}
\usepackage[export]{adjustbox}
\usepackage{subcaption}

\usepackage[show]{notes}
\newcommand{\ignore}[1]{}
\newcommand{\com}[1]{}

\newcommand{\R}{{\mathbb{R}}}
\newcommand{\GD}{GreedySpeaker}
\newcommand{\GS}{GreedySpeaker}
\newcommand{\SEM}{STEM}
\newcommand{\STEM}{STEM}

\DeclareMathOperator*{\argmaxA}{arg\,max} 

\usepackage{graphicx}   
\usepackage{tikz}       
\usepackage{multirow}   
\usepackage{bchart}     

\usepackage[super]{nth}
\usepackage{multicol} 
\usepackage{array}

\title{STEM: Unsupervised STructural EMbedding for Stance Detection}

\title{STEM: Unsupervised STructural EMbedding for Stance Detection}
\author {
    Ron Korenblum Pick\textsuperscript{\rm 1}, 
    Vladyslav Kozhukhov\textsuperscript{\rm 2},
    Dan Vilenchik\textsuperscript{\rm 2},
    Oren Tsur\textsuperscript{\rm 1}\\
}
\affiliations {
    \textsuperscript{\rm 1}Department of Software and Information Science Engineering\\
    \textsuperscript{\rm 2}Department of Communication Systems Engineering\\
    Ben Gurion University of the Negev \\
    \{ronpi,kozhukho\}@post.bgu.ac.il, \{vilenchi, orentsur\}@bgu.ac.il
}
\begin{document}

\maketitle

\begin{abstract}
Stance detection is an important task, supporting many downstream tasks such as discourse parsing and modeling the propagation of fake news, rumors, and science denial. In this paper, we propose a novel framework for stance detection.  Our framework is unsupervised and domain-independent. Given a claim and a multi-participant discussion -- we construct the interaction network from which we derive topological embedding for each speaker. These speaker embedding enjoy the following property:  speakers with the same stance tend to be represented by similar vectors, while antipodal vectors represent speakers with opposing stances. These embedding are then used to divide the speakers into stance-partitions. We evaluate our method on three different datasets from different platforms. Our method outperforms or is comparable with supervised models while providing confidence levels for its output. Furthermore, we demonstrate how the structural embedding relate to the valence expressed by the speakers. Finally, we discuss some limitations inherent to the framework.
\end{abstract}


\section{Introduction}
\label{sec:intro}
Stance detection is the task of classifying the approval level expressed by an individual toward a claim or an entity. Stance detection differs from sentiment analysis in its opaqueness. A favorable stance toward a target opinion or an entity $E$ can be expressed using a negative sentiment without any explicit mention of $E$. For example, the utterance ``I did not like the movie because of its stereotypical portrayal of the heroine as a helpless damsel in distress'' bears a negative sentiment (``I did not like...''), while one can conjecture that the speaker's stance toward feminism and women's rights is favorable. 

Understanding the stance of participants in a conversation is expected to play a crucial role in conversational discourse parsing, e.g., \cite{zakharov2021discourse}. Stance detection is used in studying  the propagation of fake news \cite{thorne2017fake,tsang2020issue}, unfounded rumors \cite{zubiaga2016stance,derczynski2017semeval}, and unsubstantiated science related to, e.g.,  global warming \cite{luo2021detecting} and the COVID-19 vaccine \cite{tyagi2020divide}.

Recent models for stance detection rely on the textual content provided by the speaker, sometimes within some social or conversational context (see Section \ref{sec:related}). These models are supervised, requiring a significant annotation effort. The dependence on language (text) as the primary, if not the sole, input, and the need for a domain (topic)-specific annotation, severely impairs the applicability of the models to broader domains and other languages \cite{hanselowski2018retrospective,xu2019adversarial}. 

Online discussions tend to unfold in a tree structure. Assuming a claim $E$ is laid at the root of the tree, each further node is a direct response to a previous node (utterance). This tree structure can be converted into an interaction network $G$, where the nodes of $G$ are speakers, and edges correspond to interactions. The edges may be weighted, reflecting the intensity of the interaction between the specific pair of speakers (see Section \ref{subsec:trees2networks}).

In this paper, we propose a novel approach for stance detection. Our method is unsupervised, domain-independent, and computationally efficient. The premise of our approach is that the conversation structure, emerging naturally in many online discussion boards and social platforms, can be used for stance detection. In fact, we postulate that the \emph{structure} of a conversation, often ignored in NLP tasks, \emph{should} be studied and leveraged within the language processing framework. 

\paragraph{Contribution} The main contribution of this paper is threefold:~(i) We introduce an efficient unsupervised and domain-independent algorithm for stance classification, based on structural speaker embedding (ii) We show how multi-agent conversational structure corresponds to speakers' stance and correlates with the valence expressed in the discussion, and (iii) The speaker embedding induces a soft classification of speakers' stances, which can be rounded to a discrete output, e.g., ``pro'', ``con'', and  ``neutral'', but can also be used to derive other interesting parameters such as the confidence level of the result, which we discuss in Section \ref{sec:discussion}.

We evaluate our model on three annotated datasets:
4forums, ConvinceMe, and CreateDebate. These datasets differ in various aspects, from the number of speakers and discussions to the variety of the topics discussed and the culture and norms shaping the conversational dynamics.  Further details about the datasets are provided in Section \ref{sec:data}. Despite these differences, our method consistently outperforms or is comparable with supervised models that were studied in other papers and were benchmarked on these datasets.

\section{Related Work}
\label{sec:related}
Stance detection gained a significant interest in recent years, e.g., \cite{somasundaran2010recognizing,walker2012stance,sridhar-etal-2015-joint,mohammad2016semeval,derczynski2017semeval,sobhani2017dataset,joseph2017constance,li2018structured,porco2020predicting,conforti2020will}, among many others. A comprehensive survey of the various settings, datasets, and computational approaches is provided in \cite{kuccuk2020stance}.

Works on stance detection differ in task specification and methodology. Broadly, stance can be assigned to an utterance or a user, and the methodology can take into account text, context or both. 

Stance at the user level, sometimes referred to as `aggregate' or `collective' stance, is addressed by \cite{murakami-raymond-2010-support,walker2012corpus,yin-etal-2012-unifying}.  A more nuanced relationship between the post and the user level is addressed by \cite{sridhar-etal-2015-joint,li2018structured,benton2018using,conforti2020will,porco2020predicting}.  We follow this observation and report results on both post and user levels. 

Modal verbs, opinion and sentiment lexicons were used in early works by \cite{somasundaran2010recognizing,murakami-raymond-2010-support,yin-etal-2012-unifying,wang-cardie-2014-improving,bar-haim-etal-2017-improving}. Recent text-based works use graphical models \cite{joseph2017constance}, CRFs \cite{hasan2013stance} and various neural architectures \cite{8272605,sun2018stance,chen-etal-2018-hybrid,kobbe2020unsupervised}, among others. These methods are language, and often domain, dependent. Unsupervised methods were also explored in the past, although to a much lesser extent than supervised ones, and using a different methodology than ours, mainly relying on topic modeling  \cite{kobbe2020unsupervised,Wei_Mao_Chen_2019}.

Leveraging the conversation structure was recently used by \cite{li2018structured,porco2020predicting} to create a global representation based on authors interaction and the text. Stance-based rumor detection is explored by \cite{wei-etal-2019-modeling}, considering the structure of the conversation, along with content. While these works leverage the conversational structure, it is done in an opaque way and is filtered through different neural architectures that combine textual queues. It is therefore hard to assess the contribution of the conversation structure to the classification task. Our framework relies solely on the structure, promoting the notion that the conversational structure is as important as the word tokens in processing conversational data.  

The intuitive assumption that consecutive 
utterances express antipodal stance is already explored by  \cite{murakami-raymond-2010-support}, using the solution to the max-cut problem to find a graph partition that reflects the stance taken by users debating policy issues in Japanese. Similarly, a solution to the max-cut problem on the \emph{conversation tree} was used by Walker et al. \shortcite{walker2012stance}.

These works are the most similar to ours, as they use the solution to the max-cut problem as the primary computational tool. Our work differs from these works in several fundamental aspects. Murakami and Raymond \shortcite{murakami-raymond-2010-support} explicitly introduce dis/agreement markers into the network representation -- agreement is coded as a positive edge weight and disagreement as a negative weight. These weights are derived from an assortment of simple heuristics and hand-crafted patterns e.g., ``I agree'', ``I disagree'', ``good point''. A fixed interpretation of these patterns overlooks cultural (or platform) norms and does not take into account nuances like irony and other discursive styles  (e.g., ``I agree with you on that point, but it is irrelevant to the issue''). Our approach does not require this noisy, culture/language-dependent and labor-intensive annotation of the network edges. Walker et al. \shortcite{walker2012stance} derive a binary output by applying a max-cut solver to the conversation tree. On the other hand, we obtain a soft classification via the speaker embedding extracted from the interaction network.

While most work on stance detection use supervised models, a number of works are unsupervised. Early works such as Somasundaran and Wiebe \shortcite{somasundaran2010recognizing} use generic opinion and sentiment lexicons. Kobbe et al. \shortcite{kobbe2020unsupervised} classify stance based on frequently used argumentation structures. Other unsupervised approaches include the use of syntactic rules for extraction of topic and aspect pairs \cite{ghosh2018unsupervised} or by extracting aspect-polarity-target information \cite{konjengbam2018debate}. These approaches are language dependant, often use external resources, and are not easily adapted to different domains and communities that present a variety of discussion norms. Our approach, however, is fully unsupervised.

Our unsupervised approach proved superior or comparable to other techniques. Moreover, the speakers' embedding allow us to derive deeper insights about the relationship between text and structure beyond the naive hypothesis that edges represent opposing stances. These insights are discussed in Section \ref{sec:discussion}.



\section{A Greedy Approach}
\label{sec:alg}

\begin{figure*}[t]
    \centering
    \includegraphics[width=0.8\textwidth]{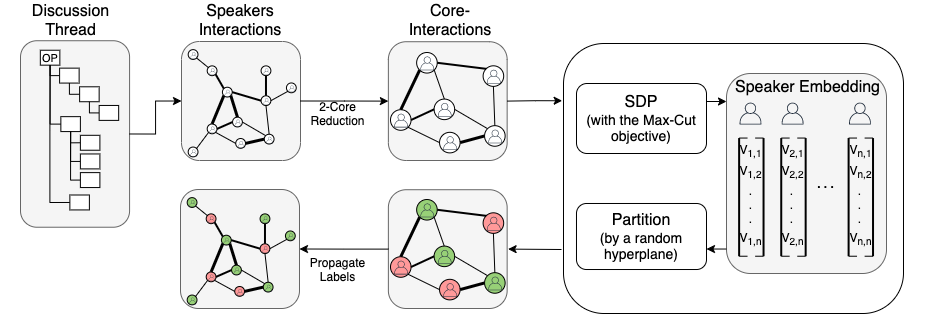}
    \caption{The workflow of \SEM. First, parsing the discussion thread (tree structure) into a weighted user-interaction graph. Then compute the 2-core of the graph. Next, run the max-cut SDP on the 2-core graph, generating the speaker embedding. A random hyperplane partitions the core speakers into two stance groups (red and green groups). Finally, propagate the labels to speakers not in the core using a simple interchanging rule.}
    \label{fig:stance-main-flow}
\end{figure*}

A naive view of the structure of an argumentative dialogue between $u$ and $v$ is that they are holding different stances. While it is tempting to assume that a simple tree structure, reflecting the turn-taking nature of a discussion, lends itself to accurate classification, this intuition does not hold for multi-participant discussions, as we demonstrate in Section \ref{sec:greedy} and the results in Section \ref{sec:experiments}. The reason is that engaging discussions tend to induce complex user interaction graphs, which are far from being bipartite. Therefore a more subtle approach is needed. We present two algorithms that build upon the same intuition. The first is a simple greedy approach and in Section \ref{sec:speakerembed} we discuss the more sophisticated method, which is based on a speaker embedding technique.



%

\subsection{From Conversation Trees to Networks}
\label{subsec:trees2networks}
A discussion could be naturally represented as a tree, where nodes correspond to posts (comment, utterance) and nodes $v_1,v_2$ are children to a parent node $r$ if they were posted, independently, as direct responses to $r$. Discussion trees capture an array of conversational patterns -- turn-taking (direct replies), the volume of direct interaction between pairs of users, and of course, the textual signal, including content and style.  However, converting the conversation tree into an interaction network may better capture the conversational dynamics. 


In the interaction network, a node corresponds to a speaker, rather than to an utterance, and an edge $e_{u,v}$ between two nodes (speakers) $u$ and $v$ indicates a direct interaction between the two. The edges can be weighted to signify the intensity of the interaction. We use the following edge weighting $w_{u,v}$:
\begin{equation}
\label{eq:InteractionEdge}
\begin{split}
  w_{u,v} = \alpha  \big( replies(u, v) + replies(v, u) \big) \\
  + \beta \big( quotes(u, v) + quotes(v, u) \big) 
  \end{split}
\end{equation}
where: $replies(u, v)$ denotes the number of times user $u$ replied to user $v$; 
$quotes(u, v)$ denotes the number of times user $u$ quoted user $v$; $\alpha$ and $\beta$ are parameters denoting the significance assigned to the corresponding interaction types (a reply or a quote). These parameters are platform-dependent and need to be adjusted to reflect the conversational norms of the target platform. For example, quoting other speakers and posts that do not directly precede an utterance are common in {\em 4forums} while scarcer in the others (see Section \ref{sec:data}). We experimented with different values to confirm robustness.




 \subsection{Algorithm 1: Greedy Speaker Labelling}
 \label{sec:greedy}
 Recall the intuitive assumption that two speakers, $u$ and $v$ that intensively engage with each other, inducing a heavy edge in the interaction network, hold opposed stances. We, therefore, begin by proposing a simple greedy algorithm based on this naive assumption.
 The algorithm receives the interaction network $G=(V,E)$ with the OP, $v_0$, marked with an abstract stance label, say $+$. In the first iteration it initializes the set of labelled speakers $S=\{v_0\}$. 
 In each consecutive iteration, it finds the heaviest edge $(u,v)$ that connects a vertex $ u \in S$ to $v \in V \setminus S$, and adds the speaker $v$ to $S$, labeling $v$ and $u$ with opposite stance labels. This algorithm is basically Prim's algorithm for minimum spanning tree, and it runs in nearly linear time, $O(|E|+|V|\log |V|)$. We call this algorithm $\GD$.
 
\section{Speaker Embedding}\label{sec:speakerembed}
A more sophisticated approach still builds upon the same intuition. It creates speaker embedding that allows a principled comparison rather than an iterative greedy assignment. A desired property of the speaker embedding, let's call it  {\em $\tau$-separability}, is that speakers with opposing stances are assigned vectors with an angle of at least $\tau$ between them (it's instructive to think of $\tau$ as close to $180^\circ$). We say that an embedding {\em $\tau$~respects the stance } if it satisfies $\tau$-separability for every pair of speakers. 

Suppose $\overrightarrow{u}$ and $\overrightarrow{v}$ are unit vectors. The separability property can be mathematically encoded by requiring that the expression in Eq.~\eqref{eq:summand} takes a larger value on pairs of opposing speakers. We use $\langle \overrightarrow{u}, \overrightarrow{v} \rangle$ for the cosine similarity between the two vectors. 

\begin{equation}\label{eq:summand}
(1-\langle \overrightarrow{u}, \overrightarrow{v} \rangle)/2  
\end{equation}

\vspace{0.5mm}

The maximal value Eq.~\eqref{eq:summand} takes is 1, which is attained if the two vectors are antipodal, namely, the angle between them is exactly $180^\circ$, and the cosine similarity is -1. Multiplying Eq.~\eqref{eq:summand} by the corresponding edge weight $w_{uv}$ ensures that the larger values are attained for relevant pairs.

Given an interaction network $G=(V,E)$, with $|V|=n$, and edge weights $w_{uv}$ for every edge $(u,v) \in E$, our goal is to find a speaker embedding $\mathcal{E}$ which respects the stance for as many speaker pairs as possible. The proposed candidate speaker embedding $\mathcal{E}$ is the solution of the optimization problem given in Eq.~\eqref{eq:SDP}, $S^n$ denoting the unit sphere in $\R^n$.

\begin{equation}\label{eq:SDP}
 \mathcal{E} =  \argmaxA_{\overrightarrow{u} \in S^n \text{ for } u \in V} \sum_{(u,v)\in E} w_{uv}\frac{1- \langle \overrightarrow{u},\overrightarrow{v} \rangle}{2}  
\end{equation}

The optimization problem in Eq.~\eqref{eq:SDP} is  a semi-definite program (SDP), and it can be solved in polynomial time using the Ellipsoid algorithm~\cite{sdpsolver}.
This SDP was suggested by \cite{Goemans1995} as a relaxation for the NP-hard max-cut problem, which is in line with our intuitive hypothesis about the nature of the interaction between speakers. Note that $n$, the dimension of the embedding, is always the number of speakers in the conversation (part of the SDP definition), unlike the tunable dimension hyper-parameter in other embedding frameworks.

\subsection{From soft to discrete classification}
The speaker embedding $ \mathcal{E}$ gives a continuous range of stance relationships, from ``total disagreement" (antipodal vectors) to ``total agreement" (aligned vectors). However, in some cases, we want to round the continuous solution to a discreet solution, say ``pro" vs. ``con". 

In addition, the separability property is relevant for pairs of speakers. Even if the embedding of every pair respects the stance, this still doesn't lend itself immediately to a partition of the {\em entire} set of speakers into two sets, ``pro" and ``con", that respects the stance. If the interaction graph is a tree, then pairwise separability immediately induces an overall consistent partition. But when cycles exist, things are messier.

We now describe how to round the speaker embedding into a partition of the speakers.
To gain intuition into the rounding technique, let's assume that the obtained embedding pairwise respects the stance, and further, that the embedding lies in a one-dimensional subspace of $\R^n$. Namely, there exists some vector $\overrightarrow{v_0}\in \R^n$ s.t. for every $u \in V$, $\overrightarrow{u} = \overrightarrow{v_0}$ or $\overrightarrow{u}= -\overrightarrow{v_0}$. In such case, the rounding is trivial: all vectors on ``one side'' are ``pro'', and all vectors on the ``other side'' are ``con'' (or vice-a-versa). 

Building upon this intuition, a random hyper-plane rounding technique is commonly used \cite{Goemans1995}. A random $(n-1)$-dimensional hyper-plane that goes through the origin is selected, and vectors are partitioned in two groups according to which side of the hyper-plane the vector lies. In the one-dimensional example, every random hyperplane will round the vectors correctly into the two opposing stance classes. More generally, the more the vectors are clustered into two ``tight'' cones, the more accurate the rounding will be (by tight, we mean that the maximum pairwise angle is small).

\begin{figure}[ht!]
\centering
\includegraphics[width=\linewidth]{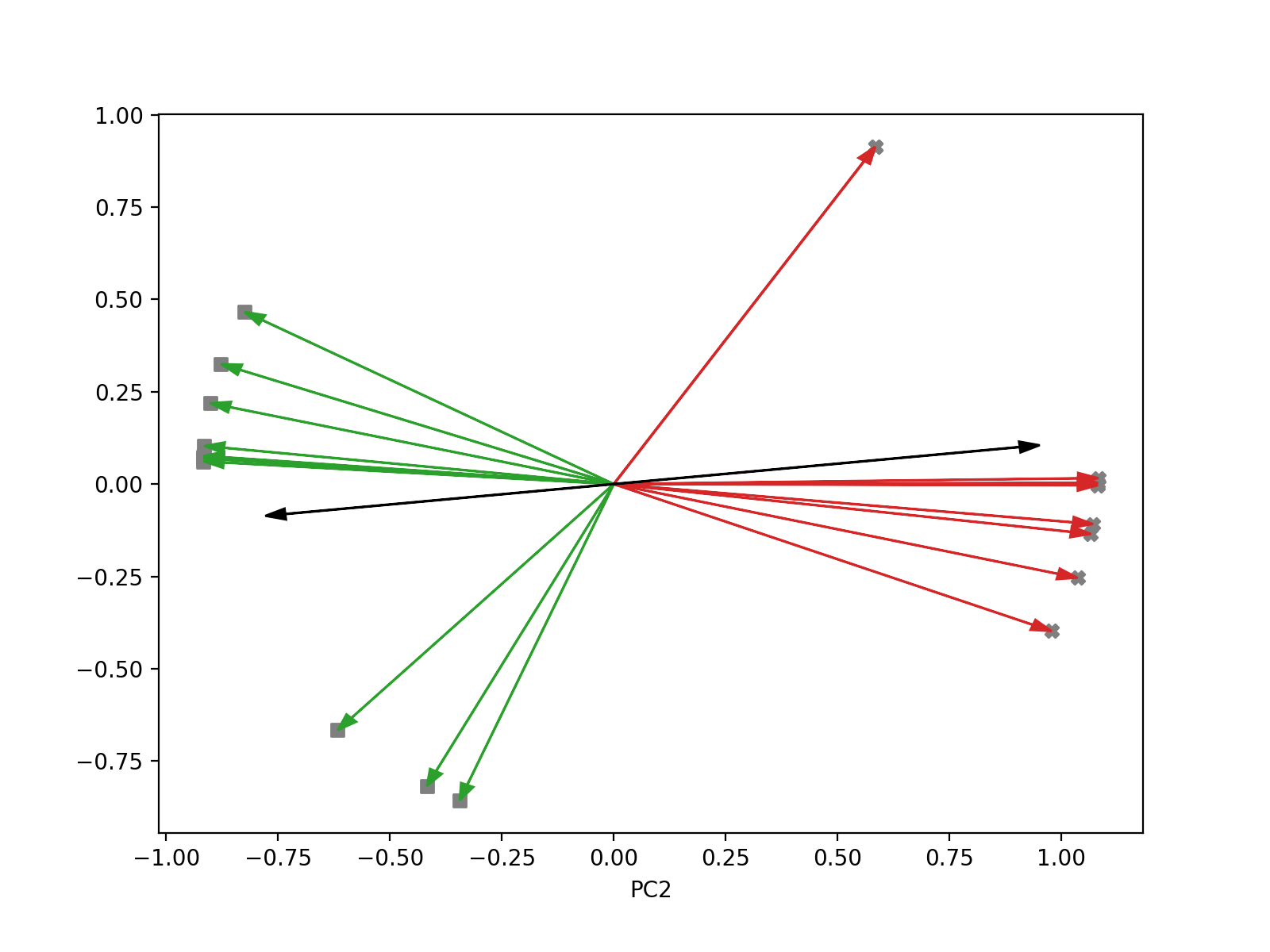}
\caption{PCA projection of the 19-dimensional speaker embedding for the core of the interaction network. Colors correspond to the speakers' labels. The black arrows to the left and right correspond to the average vector in each color class}
\label{fig:PCA_cones}
\end{figure}

\begin{figure}[ht!]
\centering
\includegraphics[width=\linewidth]{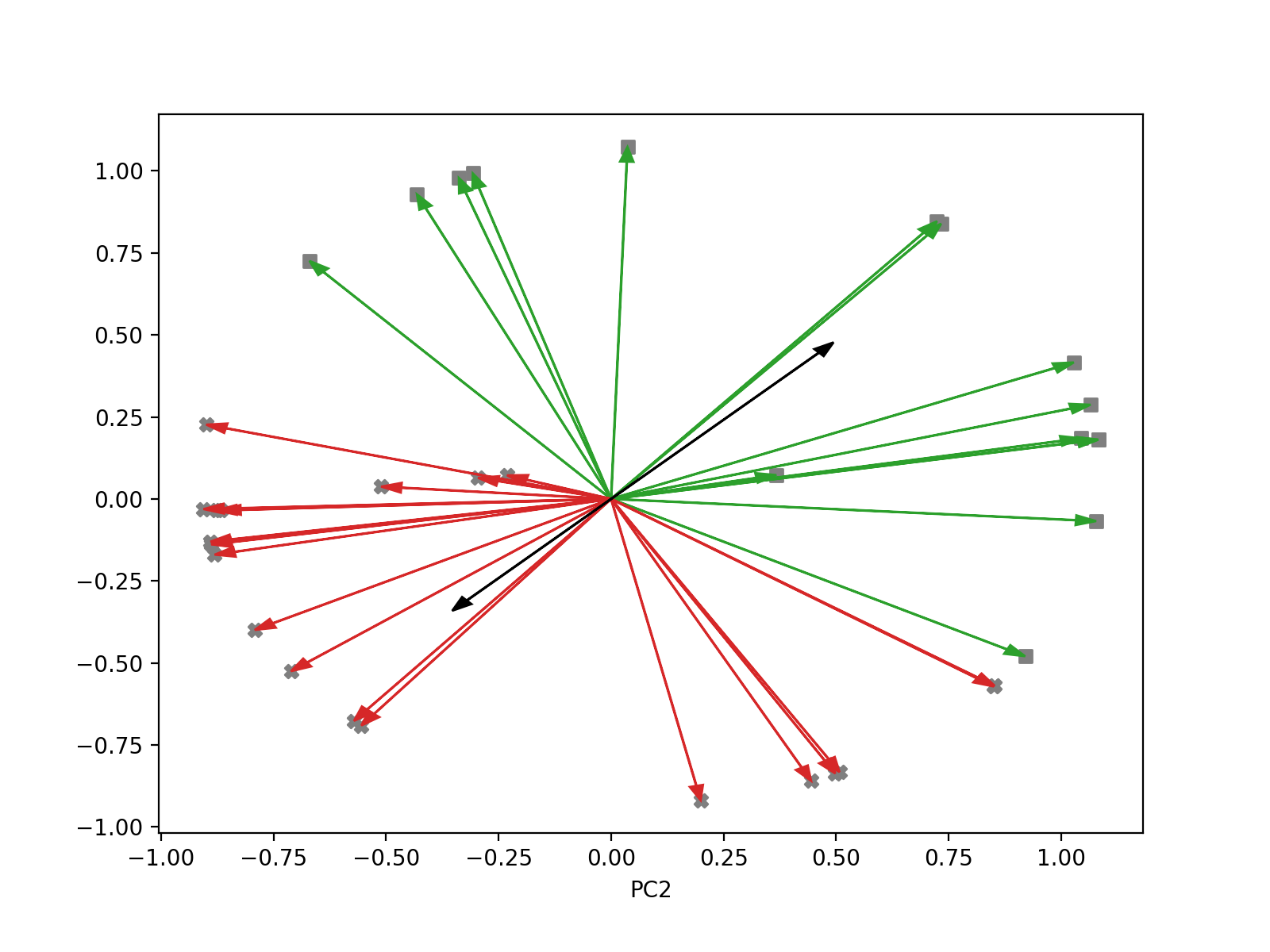}
\caption{PCA projection of the 35-dimensional speaker embedding of the core of an interaction network also from 4Forum. Shorter vectors have a larger component perpendicular to PC1 and PC2. The induced cones have a large diameter, and therefore the confidence of having a correct prediction on authors within this conversation significantly decreases. Black arrows are cone centers (again shorter).}
\label{fig:PCA_cones2}
\end{figure}

Figure \ref{fig:PCA_cones} illustrates this point: two tight cones are observed, as well as some ``straying'' vectors that are liable to wrong classification. The accuracy of the hyperplane rounding  on that conversation was 75\%. On the other hand,  Figure \ref{fig:PCA_cones2} demonstrates wider cones, and accordingly, the accuracy this time was only 64\%. Further illustration about how the  diameter of the cones corresponds to an accurate solution is given in Table \ref{table:cones_example}.

\begin{table} [!ht]
\small
\label{"tab:corr-acc-cones"}
\centering
\begin{tabular}{ | c | c c | }
\hline
 \textbf{Diameter} & \textbf{accuracy} &  \textbf{authors} \\
 \hline
 2.0 & 0.79 & 2440 \\
 1.0 & 0.80 & 2403 \\
 0.75 & 0.80 & 2341 \\
 0.5 & 0.81 & 2258 \\
 0.25 & 0.82 & 2127 \\
 0.1 & 0.83 & 1921 \\
 0.05 & 0.84 & 1761 \\
 0.01 & 0.85 & 1332 \\
 0.001 & 0.85 & 917 \\
\hline
\end{tabular}
\caption{
Accuracy of speakers classification for speakers whose vector falls inside the cone, for various cone diameters. Evidently, as the cones get tighter, the accuracy increases. The dataset used is the 4Forum conversations.}
\label{table:cones_example}

\end{table}
\subsection{Tight cones of vectors respect the stance}
It is important to note that the vectors that the SDP assigns the speakers lie in $\R^n$. This dimension provides a lot of freedom in vectors assignment (freedom which is necessary for the SDP to be solvable in polynomial time). Therefore, while the one-dimensional intuition just described is clear for a two-persons dialogue, it is not a-priori clear why the vectors in $\R^n$ should {\em simultaneously} respect the stance of all, or most, speakers in a multi-participant discussion. 

We now explore the conditions that may lead to the desired phenomenon where the SDP solution is such that the vectors are clustered in two tight cones.  These conditions are  rooted both in the network structure and in the content of the conversation.


From the perspective of the \emph{network topology}, it is easy to see that the optimal solution to Eq.~\eqref{eq:SDP} is the antipodal vectors rank-one solution we described above, where the assignment of vectors corresponds to the max-cut partition of the graph. However, crucially, Eq.~\eqref{eq:SDP} does {\em not} contain a rank constraint on the solution as this will turn the optimization problem NP-hard. Now enters the assumption that edges represent antipodal stances. If this assumption is correct, and the structure of the network is rich enough to force a unique max-cut solution, then we expect a ``tight-cones'' solution which is both aligned with the max-cut partition and with the stances.

The assumption of a unique max-cut partition may be too strong to hold for the entire graph (think for example of isolated nodes, or very sparse structures). However, for a special subgraph, the 2-core of the graph, this uniqueness may hold. Indeed, we have found that most of the SDP vectors of the speakers that belong to the 2-core of the graph (a subgraph of $G$ in which the minimal degree is 2) are arranged in a tight-cone structure. This phenomenon was observed in other papers as well, that studied related tasks such as community detection and other graph partitioning tasks \cite{PhysRevE.74.016110,Newman8577,Leskovec:2010,coloringIsEasy}.

But why should the graph contain a large 2-core in the first place? Here enters the {\em content/linguistic} aspect. We expect that captivating or stirring topics will lead to lively discussions that result in a complex conversation graph that induces a large 2-core. Together with the basic assumption that edges connect speakers with opposing stances, we arrive at the premise that in such discussions, both the SDP will produce solutions that have the tight-cones structure and that this tight-cone structure will respect the stance. Thus, when rounding the solution using the random hyper-plane technique, we expect to detect the stance of 2-cores users accurately. Section \ref{sec:discussion} elaborates on the relationship between the spirit, or valence, of the conversation and the accuracy of the algorithm.

\subsection{Algorithm 2: STEM}
\label{sec:algo2-stem}
We now formally describe our main contribution, STEM, an unsupervised structural embedding for stance detection. The below steps are also illustrated in Figure \ref{fig:stance-main-flow}. Given a conversation tree $T$, STEM operates as follows:

\begin{enumerate}
    \item Convert the conversation tree $T$ to an interaction network $G=(V,E)$, as described in Section \ref{subsec:trees2networks}.
    \item Compute the 2-core $G_C = (V_C,E_C)$ of $G$, i.e. the induced subgraph of $G$ where every node has degree at least 2 in $G_C$.
    \item Solve the SDP in Eq.~\eqref{eq:SDP} on $G_C$ to obtain a speaker embedding $\mathcal{E}$.
    \item Round the speaker embedding using a random hyper-plane. 
    \item Propagate the labels to speakers outside the core, $V \setminus V_C$, using interchanging labels assignment.
\end{enumerate}

In Step 2, we compute the core. To compute the 2-core, one iteratively removes vertices whose degree in the remaining graph is smaller than two, until no such vertex remains.

Step 5 does not lead to a contradiction since, by definition, the vertices outside the core do not induce a cycle. Therefore, the propagation of labels in the sub-graphs connected to the 2-core is consistent.

Finally, note that our algorithm produces a partition of speakers, similarly to the problem of community detection, without a label for each part (pro or con). One simple heuristic to obtain the labeling is to label the set containing the OP as ``pro''. 
Another option is to use an off-the-shelf algorithm, e.g. \cite{allaway2020zero}, and noisily label a few posts on each side before taking a majority vote.

To evaluate the performance of our algorithm without additional noise that this last step may incur, we checked the two possible ways of assigning the labels and took the one that resulted in higher accuracy. 



\section{Data}
\label{sec:data}
We evaluate our approach on three datasets: ConvinceMe \cite{anand2011cats}, 4Forums \cite{walker2012corpus}, and  CreateDebate \cite{hasan-ng-2014-taking}. These datasets were used in previous work, e.g., \cite{walker2012stance,sridhar-etal-2015-joint,abbott2016internet,li2018structured}, among others. We briefly describe each of the datasets and highlight some important aspects they differ in. A statistical description of datasets is provided in Table \ref{tab:datastats}.

\begin{table} [!t]
\small
\centering
\begin{tabular}{ | l | c |c |c| c  |}
\hline
   & 4Forums & CD & CM \\
 \hline
 \# Topics & 4 & 4 & 16  \\ 
 \# Conversations & 202 & 521 & 9,521  \\
 \# Conversations (core) & 202 & 149 & 500  \\
 \# Authors & 863 & 1,840 & 3,641  \\
 \# Authors (core) & 718 & 352 & 490   \\
 \# Posts & 24,658 & 3,679 & 42,588   \\
 \# Posts (core) & 23,810 & 1,250 & 5,876   \\
 \hline
\end{tabular}
\caption{Basic statistics of the three datasets: 4Forums, CreateDebate (CD), and Convince Me (CM). We also present the number of authors that belong to the 2-core of the interaction graph, and their posts.}
\label{tab:datastats}
\end{table}

\paragraph{ConvinceMe (CM)} ConvinceMe is a structured debate site. Speakers initiate debates by specifying a motion and stating the sides. Debaters argue for/against the motion, practically self-labeling their stance with respect to the original motion. The data was first used by Anand \shortcite{anand2011cats} and incorporated to the IAC2.0 by Abboott et al. \shortcite{abbott2016internet}.

\paragraph{4Forums} 4Forums (no longer maintained) was an online forum for political debates. It had a shallow hierarchy of topics (e.g., Economics/Tax), and discussion threads have a tree-like structure. The 4Forum stance dataset, introduced by Walker et al. \shortcite{walker2012corpus}, provides agree/disagree annotations on comment-response pairs in 202 conversations on four topics (abortion, evolution, gay marriage, and gun control). 

\paragraph{CreateDebate (CD)} Similarly to ConvinceMe, CreateDebate is a structured debate forum. Unlike ConvinceMe, the user initiating the debate does not put forward a specific assertion. Rather, she introduces an open question for the community, and speakers can respond by taking sides. 
Authors must label their posts with either a \emph{support}, \emph{clarify} or \emph{dispute} label. A collection of debates on four topics (abortion, gay rights, legalization of marijuana, Obama) was introduced by \cite{hasan-ng-2014-taking}. This dataset contains many degenerate conversations -- speakers responding to the prompt question without engaging in a conversation with other speakers. We filtered out these degenerate conversations, keeping 541 conversation trees (see Table \ref{tab:datastats}). The root of each of the trees is an original response to the initial questions.



\section{Evaluation}
\label{sec:experiments}

\begin{table*}[ht!]
\small
\centering
\begin{tabular}{ | c | c c c c | c | }
\hline
 \textbf{Model} & Abortion & Gay Rights & Marijuana & Obama & Average \\
 \hline
 PSL (Sridhar et al., 2015) & 0.67 & 0.73 & 0.69 & 0.64 & 0.68 (macro)  \\
 Global Embedding (Li et al., 2018) & 0.81 & 0.77 & 0.77 & 0.65 & 0.75 (macro) \\
 \hline
$\GS$ (full) & 0.80 & 0.81 & 0.74 & 0.79 & 0.79  \\
$\STEM$ (core) & 0.91 & 0.82 & 0.82 & 0.82 & 0.86  \\
$\STEM$ (full) & 0.90 & 0.85 & 0.74 & 0.86 & 0.86  \\
\hline
\end{tabular}
\captionsetup{justification=centering}
\caption{Average accuracy on posts' stance classification of \textbf{CreateDebate} discussions.}
\label{table:CD_1}
\end{table*}

\begin{table*} [ht!]
\small
\centering
\begin{tabular}{ | c | c c c c | c | }
\hline
 \textbf{Model} & Abortion & Gay Rights & Marijuana & Obama & Average \\
 \hline
 PSL (Sridhar et al., 2015) & 0.67 & 0.74 & 0.75 & 0.63 & 0.71 (macro) \\
 \hline
 $\GD$ (full) & 0.87 & 0.86 & 0.76 & 0.85 & 0.85  \\
 $\SEM$ (core) & 0.91 & 0.79 & 0.86 & 0.83 & 0.88  \\
 $\SEM$ (full) & 0.86 & 0.80 & 0.70 & 0.83 & 0.85  \\ 
\hline
\end{tabular} 
\captionsetup{justification=centering}
\caption{Average accuracy for authors' stance classification for \textbf{CreateDebate} discussion.}\label{table:CD_2}

\end{table*}

\begin{table*} [ht!]
\small
\label{table:results-4forum-posts}
\centering
\begin{tabular}{ | c | c c c c | c | }
\hline
 \textbf{Model} & Abortion & Evolution & Gay Marriage & Gun Control & Average \\
 \hline
 PSL (Sridhar et al., 2015) & 0.77 & 0.80 & 0.81 & 0.69 & 0.77 (macro) \\
 Global Embedding (Li et al., 2018) & 0.87 & 0.82 & 0.88 & 0.83 & 0.85 (macro) \\
 \hline
 $\GS$ (full) & 0.62 & 0.61 & 0.60 & 0.63 & 0.62 \\
 $\STEM$ (core) & 0.93 & 0.88 & 0.89 & 0.85 & 0.89 \\
 $\STEM$ (full) & 0.92 & 0.87 & 0.88 & 0.84 & 0.89 \\
\hline
\end{tabular}
\captionsetup{justification=centering}
\caption{Average accuracy on posts' stance classification of \textbf{4Forum} discussions.}\label{table:4F_1}
\end{table*}

\begin{table*}[ht!]
\small
\label{"tab:results-4forum-authors"}
\centering
\begin{tabular}{ | c | c c c c | c | }
\hline
 \textbf{Model} & Abortion & Evolution & Gay Marriage & Gun Control & Average \\
 \hline
 PSL (Sridhar et al., 2015) & 0.66 & 0.79 & 0.77 & 0.68 & 0.73 (macro)  \\
 \hline
 $\GD$ (full) & 0.61 & 0.59 & 0.59 & 0.62 & 0.60 \\
 $\SEM$ (core) & 0.84 & 0.78 & 0.79 & 0.74 & 0.79 \\
 $\SEM$ (full) & 0.79 & 0.75 & 0.77 & 0.71 & 0.76 \\
\hline
\end{tabular}
\captionsetup{justification=centering}
\caption{Average accuracy of authors' stance classification for \textbf{4Forum} discussions.}
\label{table:4F_2}
\end{table*}

\begin{table} [!ht]
\small
\centering
\begin{tabular}{ | c | c c c | }
\hline
 \textbf{Topic} & \# Posts & \SEM & Walker \\
\hline
 Gay Marriage & 708 & 0.98 & 0.84 \\
 Evolution & 688 & 0.99 & 0.82 \\
 Communism Vs Capitalism & 185 & 0.99 & 0.70 \\
 Marijuana Legalization & 261 & 0.98 & 0.73 \\
 Gun Control & 314 & 0.95 & 0.63 \\
 Abortion & 834 & 0.96 & 0.82 \\
 Climate Change & 255 & 1.00 & 0.64 \\
 Israel/Palestine & 36 & 1.00 & 0.85 \\
 Existence Of God & 842 & 0.98 & 0.75 \\
 Immigration & 166 & 0.87 & 0.67 \\
 Death Penalty & 474 & 0.98 & 0.65 \\
 Legalized Prostitution & 108 & 0.88 & NA \\
 Vegetarianism & 43 & 1.00 & NA \\
 Women In The Military & 22 & 1.00 & NA \\
 Minimum Wage & 14 & 0.95 & NA \\
 Obamacare & 101 & 0.98 & NA \\
 Other & 37,537 & 0.95 & NA \\
\hline
\end{tabular}
\caption{Average accuracy of post-level stance achieved by  $\STEM$ and the Max-Cut algorithm from  \cite{walker2012stance} on the {\bf ConvinceMe} dataset.}
\label{table:CM_1}.
\end{table}

 \ignore{

 }

{\bf Implementation} Our approach uses only two hyper-parameters, $\alpha$ (reply weight) and $\beta$ (quote weight), which are used to compute the weights of the edges in the interaction graph, see Eq.~\eqref{eq:InteractionEdge}. The optimal values may differ between datasets, as the conversational norms may differ. We fixed the values manually; for {\em 4Forum} we used $\alpha=0.02, \beta=1.0$ as participants tend to reply to the OP regardless of the content to which they are replying, and only quote the relevant content instead. For {\em CreateDebate} and {\em ConvinceMe} we used $\alpha=1.0, \beta=0.0$ as quotes rarely  used. 

To solve the SDP optimization in Eq.~\eqref{eq:SDP} we used standard open-source code libraries, PICOS \footnote{https://picos-api.gitlab.io/picos/index.html} and CVXOPT \footnote{https://cvxopt.org}. All the source code required for conducting the experiments and reproducing our results is available on Github\footnote{https://github.com/NasLabBgu/STEM} (including the random seed). The average running-time for computing the solution for a single conversation (including the SDP) was $0.41$ seconds. The average time was taken over $202$ conversations from {\em 4Forums} as this datasets contains the largest conversations, with an average of $15$ speakers in the core-graph ($52$ speakers max). We ran the experiment on a machine equipped with a processor with 8 cores and {\tt 16GB} RAM (we didn't use a GPU for the computation).

\vspace{1.5mm}

\noindent{\bf Evaluation} We evaluated $\GD$ (Section \ref{sec:greedy}) and $\SEM$ (Section \ref{sec:algo2-stem}) on the three datasets described in Section \ref{sec:data}, both at the speaker level and  the post level. The 4Forum dataset had both post-level and speaker-level labels. 

In cases where ground-truth labels were available only at the post level (CD and CM), we extended the post-level labeling to speaker-level by taking a majority vote over the posts of each user; in cases where the results were reported at the post level (CD and 4Forum), we labeled the posts according to the stance of that speaker.

Results, compared to previous work on the CreateDebate dataset, are presented in Tables \ref{table:CD_1} and \ref{table:CD_2}. Two types of results are reported: the accuracy of each algorithm on the speakers that belong to the 2-core, and the accuracy over all speakers. The results are given at the post level (Table \ref{table:CD_1}) and speaker level (Table \ref{table:CD_2}). Similar results on the 4Forums dataset are presented in Tables \ref{table:4F_1} and \ref{table:4F_2}.

As evident from the tables, $\SEM$ outperforms other approaches across all topics and datasets.  Also evident from the tables is that the accuracy of $\SEM$ on the 2-core is always higher than the accuracy, over all speakers. We note that even the $\GD$ algorithm significantly outperforms SOTA results reported in the literature. 

We complete our evaluation with a direct comparison to the Max-Cut approach used by Walker et al. \shortcite{walker2012stance}. Walker et al. solve the Max-Cut problem on the conversation tree (where posts are also linked to authors), using some Max-Cut solver (not SDP). They report results at the post level for the ConvinceMe dataset. Table \ref{table:CM_1} presents results for each topic separately, demonstrating the usefulness of our more elaborate way of using the Max-Cut intuition.


\section{Discussion}
\label{sec:discussion}

\paragraph{Valence}
Our work suggests that a rich interaction graph structure leads to useful speaker embedding. The latent link between the linguistic aspects of the conversation and the graph structure may relate to the valence of the conversation. To explore this, we computed the valence of the conversations in 4Forum using Python's \texttt{PySentiStr} \cite{sentimentPython}. Each conversation was scored with the average valence of its posts. We found that the average accuracy of $\STEM$ on conversations whose valence is at the lower end (0--0.5), was 0.75, while the average accuracy on conversations with medium valence (0.5--0.8) was 0.8, and the average accuracy on conversations exhibiting high valence (0.8--1) is increased to 0.92. These results support our hypothesis that stirred-up discussions lead to richer interaction graph structure, resulting in more accurate speaker embedding. Future work should further investigate this link between content, stance, and conversation structure.

\paragraph{Confidence} The soft classification induced by the speaker embedding allows us to attribute confidence levels to our result. Specifically, Table \ref{table:cones_example} demonstrates how the accuracy of the algorithm improves as we perform the rounding of vectors on increasingly tighter cones. Therefore along with the binary stance classification, we can add a score, which is proportional to the cone diameter of the 2-core, which informs the user how certain we are about the accuracy of our results. This is illustrated in Figure \ref{fig:PCA_cones2}, where a larger diameter of the cones resulted in lower accuracy, 64\%.


Rounding the embedding of the 2-core and propagating the results to the non-core speakers may be sub-optimal. As Table \ref{table:cones_example} suggests, it might be better to round a subgraph of the 2-core that corresponds to tighter cones, at the expense of labeling fewer speakers in the rounding step, and then propagate the labels to the remaining core and non-core vertices. 

\paragraph{Limitations} Finally, let us mention the limitations of our approach. The task of stance classification is not limited to structured platforms like ConvinceMe or 4Forum. Indeed, debates take place on general-purpose platforms such as Twitter or Facebook, where a wider range of reactions is available. We have not tested our method on such data, and it may be the case that the conversational norms on these platforms differ radically from those in the three datasets we used. 

Another limitation is the 2-core requirement. It might be that discussions in some platforms result in core-free graphs or graphs with several small 2-cores. We have tested our method on interaction graphs that are trees. Our approach worked well for some trees while it stumbled on others. 


\section{Conclusion}\label{sec:conclusion}
We proposed an unsupervised and domain-independent approach to stance detection. Our approach leverages the conversation structure to compute a useful speaker embedding. We demonstrate the benefits of this approach by evaluating it on three datasets and comparing the performance to the state-of-the-art results reported on them. Moreover, we have demonstrated how the speaker embedding allows for soft classification, which can be viewed as a confidence measure for classification results of specific instances. Finally, we explore the relations between the valence expressed in a discussion, the conversational structure, the interaction network, and the participants' stance. We observed a correlation between stance classification accuracy and the valence levels, as well as a correlation between the accuracy and the size of the network core. These relations will be explored in future work.

\bibliography{stance}

\end{document}